\def\year{2019}\relax
\documentclass[letterpaper]{article} 
\usepackage{aaai19}  
\usepackage{times}  
\usepackage{helvet}  
\usepackage{courier}  
\usepackage{url}  
\usepackage{graphicx}  
\usepackage{xcolor}
\usepackage{amsmath,amssymb} 
\usepackage{caption}
\usepackage{multirow}
\def\real{\mathbb{R}}

\begin{document}
%
\title{Mono3D++: Monocular 3D Vehicle Detection \\with Two-Scale 3D Hypotheses and Task Priors}
\author{Tong He$^{\dagger,\ddagger}$,\ \ Stefano Soatto$^{\dagger}$\\
$^{\dagger}$University of California, Los Angeles\ \ \ \ $^{\ddagger}$Megvii (Face++)\\
\{simpleig, soatto\}@cs.ucla.edu\\
}
\maketitle

\begin{abstract}
We present a method to infer 3D pose and shape of vehicles from a single image. To tackle this ill-posed problem, we optimize two-scale projection consistency between the generated 3D hypotheses and their 2D pseudo-measurements. Specifically, we use a morphable wireframe model to generate a fine-scaled representation of vehicle shape and pose. To reduce its sensitivity to 2D landmarks, we jointly model the 3D bounding box as a coarse representation which improves robustness. We also integrate three task priors, including unsupervised monocular depth, a ground plane constraint as well as vehicle shape priors, with forward projection errors into an overall energy function. 
\end{abstract}

\section{Introduction}
\noindent Objects are regions of three-dimensional (3D) space that can move independently as a whole and have both geometric and semantic attributes (shapes, identities, affordances, etc.) in the context of a task. In this paper, we focus on vehicle objects in driving scenarios. 
Given an image, we wish to produce a posterior probability of vehicle attributes in 3D, or at least some point-estimates from it.

Inferring 3D vehicles from a single image is an ill-posed problem since object attributes exist in 3D but single images can only provide partial pseudo-measurements in 2D. Therefore, we propose to solve this task by tackling two issues: (i) how to ensure 3D-2D consistency between the generated 3D vehicle hypotheses and their corresponding 2D pseudo-measurements, which requires strong 3D hypotheses generators as well as robust scoring mechanisms; (ii) how to refine 3D hypotheses with task priors that can be integrated into an easy-to-optimize loss function.
 
\begin{figure}
\centering
\resizebox{1 \columnwidth}{!}{

\includegraphics[height=6.5cm]{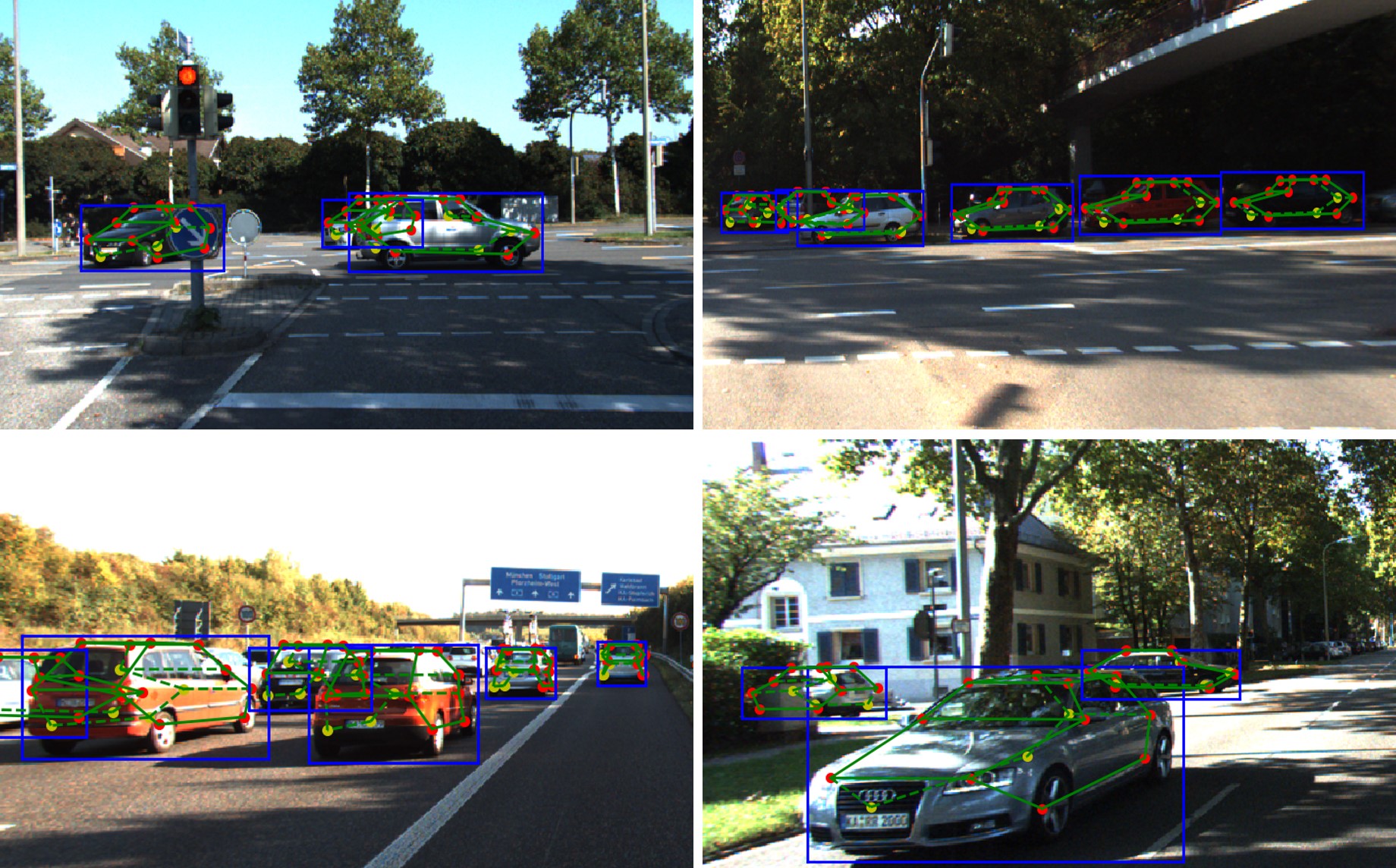}

}
\caption{Representative 3D detection results, shown as projections on the input images. 3D morphable shape models of each vehicle are colored in green. Dashed green lines are occluded edges. For the 14 vertices of each morphable shape model, visible vertices are colored in red and occluded ones in yellow. 2D bounding boxes are colored in blue.}
\label{visual_output}
\end{figure}

\begin{figure*}
\centering
\resizebox{1 \textwidth}{!}{

\includegraphics[height=6.5cm]{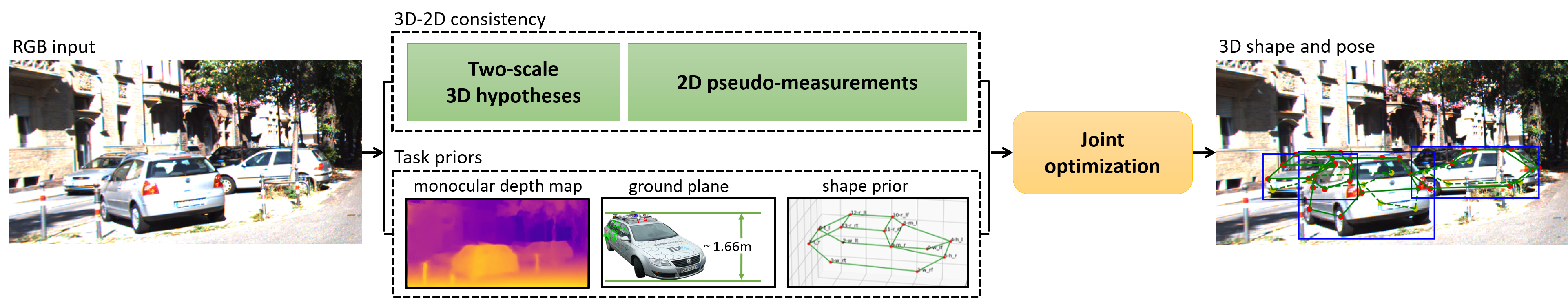}

}
\caption{Our system takes a single image as input, and generates vehicles' 3D shape and pose estimation in camera coordinates.}
\label{pipeline_0}
\end{figure*}

For the first problem, we use a joint modeling method that leverages two different 3D hypotheses generation schemes: one serves as a coarse representation of vehicle shape and pose while the other is fine-scaled. We design end-to-end trained deep networks to generate 3D hypotheses for each vehicle instance in the form of both 3D bounding box and morphable wireframe model (a.k.a linear shape model). Shape and pose parameters will be adjusted according to the 2D pseudo-measurements via an optimization approach. A wireframe model can determine shape and pose more precisely than a 3D bounding box, but it is very sensitive to the 2D landmark measurements which can be easily affected by issues like partial occlusions, shadow, low resolution, etc. Therefore, we jointly model the 3D bounding box projection constraint to improve its robustness. 
We conduct ablation studies to demonstrate benefits brought by jointly modeling the coarse and the fine-scaled 3D object pose and shape representations.

For the second problem, we consider three constraints on vehicles. Cars should stay on the ground plane, should look like a car, and should be at a reasonable distance from the observation camera. The first argument serves as a supporting plane constraint for vehicles. The second argument is a prior term for vehicle shapes. The last argument indicates that vehicle translation in camera coordinates should be constrained by a monocular range map of the current driving scene. These constraints are jointly modeled with 3D-2D consistency terms in order to further improve vehicle shape and pose estimation. 

In summary, in this paper we propose an approach for vehicle 3D shape and pose estimation from a single image that leverages the coarse and the fine-scaled 3D hypotheses, as well as multiple task priors, as shown in Fig.~\ref{pipeline_0}, for joint inference. Our contributions are:
\begin{itemize}
\item We propose and empirically validate the joint modeling of vehicles' coarse and fine-scaled shape and pose representations via two complementary 3D hypotheses generators, namely 3D bounding box and morphable wireframe model.

\item In our method, we model multiple priors of the 3D vehicle detection task into easy-to-optimize loss functions. Such priors include unsupervised monocular depth, a ground plane constraint and vehicle shape priors.

\item We build an overall energy function that integrates the proposed two improvements which  
improves the state-of-the-art monocular 3D vehicle detectors on the KITTI dataset. 
\end{itemize}

\section{Related work}
To produce rich descriptions of vehicles in 3D space, methods leveraging various data are proposed, such as video, RGB-D, or RGB, among which RGB methods are most related to our work. 

\subsubsection{Video methods:} In~\cite{song2015joint,murthy2017shape,falak_icra_2016,Dong2017VisualInertialSemanticSR,Fei2018VisualInertialOD}, temporal information is explored for (moving) 3D objects localization by a recursive Bayesian inference scheme or optimizing loss functions that are based on non-rigid structure-from-motion methods~\cite{torresani2004learning}.

\subsubsection{RGB-D methods:} MV3D~\cite{chen2017multi} encodes lidar point clouds into multi-view feature maps, which are fused with images, and uses 2D convolutions for 3D localization. In contrast, F-PointNet~\cite{qi2017frustum} directly processes lidar point clouds in 3D space using two variants of PointNet~\cite{qi2017pointnet++} for 3D object segmentation and amodal detection. Other methods that also use lidar point clouds include~\cite{ren20183d,xu2017pointfusion}. 3DOP~\cite{3dopNIPS15} exploits stereo point clouds and evaluates 3D proposals via depth-based potentials.

\subsubsection{RGB methods:} Mono3D~\cite{chen2016monocular} scores 3D bounding boxes generated from monocular images, using a ground plane prior and 2D cues such as segmentation masks. Deep3DBox~\cite{mousavian20173d} recovers 3D pose by minimizing the reprojection error between the 3D box and the detected 2D bounding box of the vehicle. Task priors are not jointly modeled with 3D-2D innovation terms. 3DVP~\cite{xiang2015data} proposes 3D voxel with occlusion patterns and uses a set of ACF detector for 2D detection and 3D pose estimation. Its follow-up work, SubCNN~\cite{xiang2017subcategory}, uses deep networks to replace the ACF detectors for view-point dependent subcategory classification. Active shape models are explored in~\cite{zia2011revisiting,zia2013explicit,zia2014cars,zia2015towards} for vehicle modeling. CAD models are rendered in~\cite{mottaghi_cvpr15,choy2015enriching} for 3D detection by hypotheses sampling/test approaches using image features, such as HOG~\cite{dalal2005histograms}. In the state-of-the-art DeepMANTA~\cite{chabot2017deep}, vehicle pose is adjusted by 3D-2D landmark matching. These approaches only model either the coarse or the fine-scaled 3D shape and pose representation of a vehicle, thus have limitations in accuracy and robustness. Moreover, task priors, such as monocular depth of the current driving scene, vehicle shape priors, supporting plane constraints, etc., are only partly considered and not jointly optimized with forward projection errors.

\section{Method}
We wish to infer the posterior distribution of object pose $g \in SE(3)$ and shape $S \subset {\mathbb R}^3$, given an image $I$, $P(g, S | I)$, where $SE(3)$ denotes the Euclidean group of rigid motions that characterize the position and orientation of the vehicle relative to the camera, and shape $S$ is characterized parametrically for instance using a point cloud or a linear morphable model. In the Supplementary Material \footnote{https://tonghehehe.com/det3d} we describe all the modeling assumptions needed to arrive at a tractable approximation of the posterior, maximizing which is equivalent to minimizing the weighted sum:
\begin{equation}
\resizebox{1.0\hsize}{!}{$E(g, S) = E_{\rm 2D3D} + \lambda_1 E_{\rm LP} + \lambda_2 E_{\rm MD}+ \lambda_3 E_{\rm GP} + \lambda_4 E_{\rm S}$}
\label{sum_energy}
\end{equation}
in which the first two terms indicate forward projection errors of the coarse and the fine-scaled 3D hypotheses. The last three terms, respectively, represent constraints enforced via unsupervised monocular depth, a ground plane assumption, and vehicle shape priors. In the next few sections we formalize and introduce details of our inference scheme, as reflected in the joint energy function~\eqref{sum_energy}, and how we generate the 3D hypotheses as well as the 2D pseudo-measurements via deep networks to facilitate its optimization.

\subsection{Notation}
We assume we are given a color image $I: D \subset {\mathbb R}^2 \rightarrow {\mathbb S}^2$ sampled as a positive-valued matrix. An axis-aligned subset $b \subset D$ is called ``2D bounding box'', {and represented by the location of its center $(t_x, t_y) \in \real^2$, and scales $(e^w, e^h) \in \real_+^2$, all in pixel units and represented in exponential coordinates $(w, h)$ to ensure positivity.} We assume the camera is calibrated so these can be converted to Euclidean coordinates. Equivalently, a 2D bounding box can be represented by an element of the scaled translation subgroup of the affine group in 2D:
\begin{equation}
g_b = \left[\begin{array}{ccc} e^w & & t_x \\ & e^h & t_y \\ & & 1 \end{array}\right]\in {\mathbb{A}(2)}.
\end{equation}

In space, we call a gravity-aligned parallelepiped $B \subset \real^3$ a 3D bounding box, resting on the ground plane, whose shape is represented by three scales $\sigma = (e^L, e^H, e^W) \in \real_+^3$, again in exponential coordinates to ensure positivity, and whose pose is represented by its orientation $\theta \in [0, 2\pi)$ and position on the ground plane $T \in \real^3$ relative to the camera reference frame. A 3D bounding box can also be represented as an element of the scaled translation subgroup of the affine group in 3D:
\begin{equation}
g_B = [R(\theta), \ e_4] \left[\begin{array}{cccc} e^L & &   & T_X \\ & e^H & & T_Y\\ & & e^W & T_Z \\ & & & 1 \end{array}\right]\in {\mathbb{A}(3)}
\label{eq-B}
\end{equation}
where $R(\theta)$ is a rotation around $Y$ by $\theta$, so $(R(\theta), T) \in SE(3)$, and $e_4^T = [0, \ 0, \ 0, \ 1]$. Here $T_Y = 0$ is the camera height from the ground plane. Assuming the ground plane is represented by its (scaled) normal vector $N \in \real^3$, we describe it as the locus $\{ T \in \real^3 \ | \ N^T T = 1\}$ (the ground plane cannot go through the optical center). Therefore, the vector $T$ is subject to the constraint $N^T T = 1$.

We call $S \subset \real^{3}$ a shape, represented by a set of $K$ points\footnote{We overload the notation and use $P$ for for points in space and probabilities. Which is which should be clear from the context.} $P_k \in \real^3$ in a normalized reference frame. Note that $p\in D \subset \real^2$ are corresponding landmark points within the 2D bounding box. Equivalently, $S \in \real^{3\times K}/{\mathbb{A}(3)}$ is in the affine shape space~\cite{kendall1984shape} of $K$ points, where the quotient is restricted to transformations of the form \eqref{eq-B}. $Z: D \subset \real^2 \rightarrow \real_+$ is a depth map, that associates to each pixel $(x, y) \in D$ a positive scalar $Z(x, y)$.

\subsection{Inference scheme}
As shown in Fig.~\ref{pipeline_0}, the inference criterion we use combines a generative component, whereby we jointly optimize the innovation (forward prediction error) between the projection of the 3D hypotheses and the image pseudo-measurements, monocular depth map constraints, geometric constraints (ground plane), in addition to penalizing large deformations of the shape prior. In the Supplementary Material we derive an approximation of the posterior of 3D pose and shape $P(g_B,S|I)$, maximizing which is equivalent to minimizing the negative log:
\begin{equation}
-\log[P(g_b|g_B)P(p|g_B,S)P(Z_b|T_Z)P(T)P(S)]
\end{equation}
which is in accordance with~\eqref{sum_energy}. The first term is the 2D bounding box compatibility with the projected 3D bounding box. It is a coarse statistic of the geometric innovation.
\begin{equation} \label{coarse_geo}
E_{\rm 2D3D} = \| g_b - \pi(g_B) \|
\end{equation}
where the subscript suggests 2D/3D consistency, and $\pi$ denotes the central perspective map, assuming a calibrated camera. The second term is the fine-scaled geometric innovation, {\em i.e.}, the distance between the predicted position of projected wireframe model vertices, and their 2D pseudo-measurements by a landmark detection network
\begin{equation} \label{fine_geo}
E_{\rm LP} = \sum_{k=1}^K \| p_k - \pi( g_B P_k) \|_2^2
\end{equation}
where LP stands for landmark projection. To approximate the third term, we produce a dense range map of the current driving scene via an unsupervised monocular depth map estimation network.
\begin{equation}
E_{\rm MD} = \|T_Z - Z_b\|
\end{equation}
where MD means monocular depth, and $Z_b \in \real$ is the average depth of an image crop specified by $(I,g_b): D \subset {\mathbb R}^2 \rightarrow {\mathbb S}^2$. The fourth term assumes a geometric constraint by the ground plane, characterized by the normal vector $N$.
\begin{equation}
E_{\rm GP} = \| N^{T}T - 1\|
\end{equation}
in which GP indicates ground plane. As generic regularizers, we also assume small deformations (shape coefficients $\alpha_n$ close to their mean) of the shape model.
\begin{equation}
E_{\rm S} = \sum_{n=1}^{N}\| \alpha_{n} - \frac{1}{N} \sum_n \alpha_{n} \|_2^2
\end{equation}

The overall loss function is the weighted sum, with multipliers $\lambda$:
\begin{align} \label{joint_energy}
\begin{split}
E =\ & E_{\rm 2D3D}(g_b, g_B) + \lambda_1 E_{\rm LP}(p, g_B, P) \\
    & + \lambda_2 E_{\rm MD}(Z_b,T)+ \lambda_3 E_{\rm GP}(T) + \lambda_4 E_{\rm S}(\alpha)
\end{split}
\end{align}

\subsection{Mono3D++ network}
\begin{figure}[h]
\centering
\resizebox{1 \columnwidth}{!}{

\includegraphics[height=6.5cm]{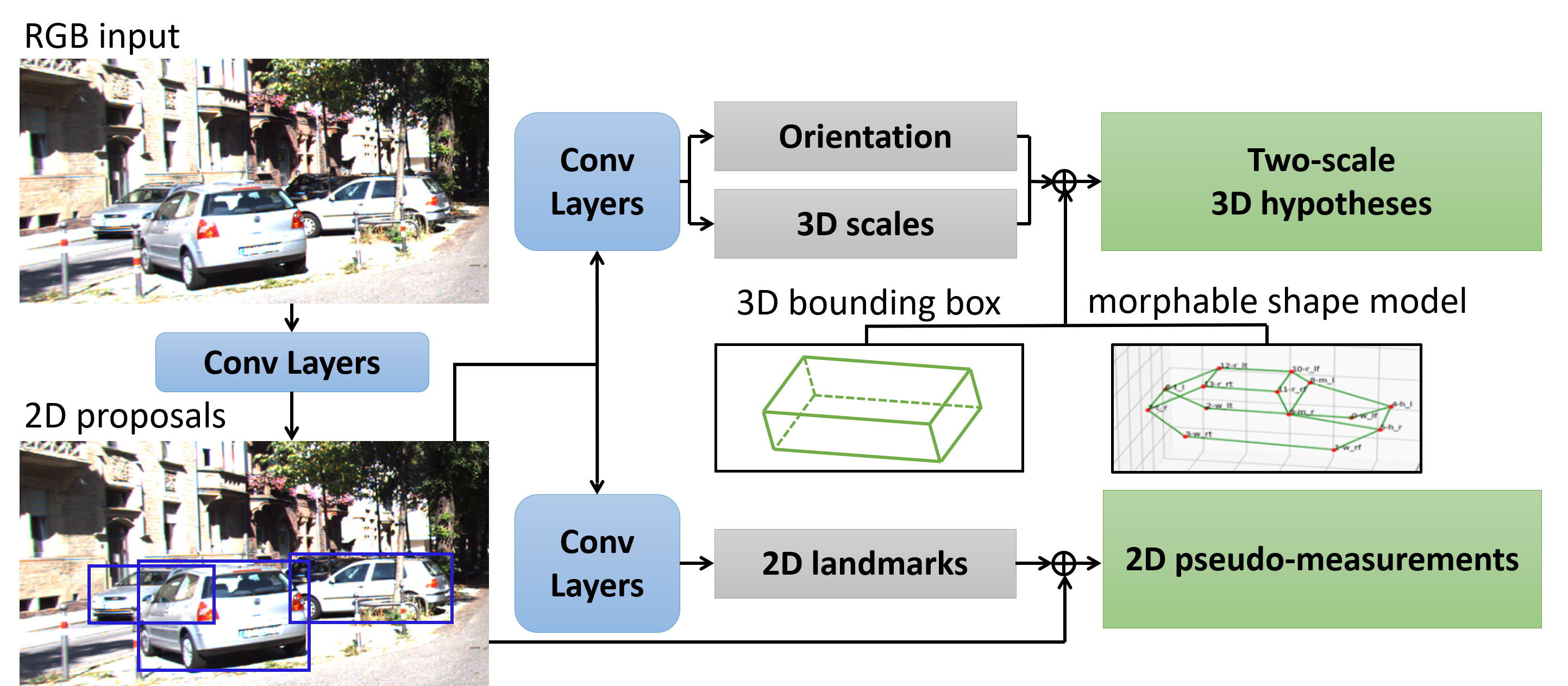}

}
\caption{The two-scale 3D hypotheses consist of the rotated and scaled 3D bounding box and morphable wireframe model. The image pseudo-measurements include 2D bounding boxes and landmarks. In our inference scheme, we use the hypotheses and the pseudo-measurements to initialize the optimization of ~\eqref{joint_energy} and generate the final 3D pose and shape estimation of a vehicle.}
\label{pipeline_1}
\end{figure}

Our inference scheme leverages independence assumptions to factor the posterior probability into components, resulting in the compound loss described above. To initialize the minimization of~\eqref{joint_energy}, we separate it into modules that are implemented as deep networks, which is shown in Fig.~\ref{pipeline_1}. In the next paragraphs we describe each component in more details.

\subsubsection{2D Bounding box.} We use a one-stage detection architecture similar to SSD~\cite{liu2016ssd} to estimate a posterior over (a regular subsampling of) the set of bounding boxes, $P(l, g_b|I)$. Before setting putative bounding boxes on latent feature maps at different scales, we fuse feature maps from shallow and deep layers. It is shown in~\cite{ren2017accurate} that this step can improve both the semantic meaning and the resolution of the feature maps for better object detection.
\begin{align} \label{loss_2D_box}
& H_{P,Q}(l_j, \hat l_i) + \lambda  d(g_{b_j}, \hat g_{b_i} )\chi(l_j)
\end{align}
where $j = j(i) = \arg\max_j {\rm IoU}(b_j, \hat b_i)$. Here $H$ denotes the cross-entropy between the true class posterior $P(l | I)$ and the model realized by our network $Q(\hat l | I)$, and $d$ is a distance in ${\mathbb A}(2)$ that sums the Euclidean distance of the translational terms and the exponential coordinates of the scale terms, $(w, h)$. Here $i$ is the index of each putative bounding box (a.k.a. ``anchor box'') which is sampled with negative/positive sample ratio of 3:1 from a larger group of regularly sampled anchor boxes on multiple latent feature maps. The index $j$ to be chosen to match $i$ consists of a data association problem~\cite{bowman2017probabilistic}. We apply an indicator function $\chi(\cdot)$ before the bounding box coordinate regression term so that this loss is only optimized for positive anchor boxes.

\subsubsection{2D Landmark.} We employ a stacked hourglass network~\cite{newell2016stacked}, with skip-connections, to approximate the posterior of individual landmarks within each 2D bounding box $P(p_k|I,\hat g_b)$. We use the mean-square error as loss.
\begin{align} \label{loss_landmark}
& \frac{1}{K}\sum_{k=1}^{K}||\hat{w}_{k,i}-w_{k,i}||^{2}
\end{align}
where $\hat{w}_{k,i} \in \mathbb{R}^{64 \times 64}$ is the predicted heat map for the $k_{th}$ landmark and $w_{k,i} \in \mathbb{R}^{64 \times 64}$ is a 2D Gaussian with standard deviation of 1 pixel centered at the $k_{th}$ landmark ground truth location. When a landmark is occluded or out of view, its ground truth heat map is set to all zeros. 
Each vehicle is modeled by 14 landmarks.

\subsubsection{3D Orientation and scale hypotheses.} $P(\theta,\sigma|I,\hat g_b)$ is approximated by a deep network with ResNet backbone~\cite{he2016deep}, yielding $Q(\hat g_B | I, \hat g_b)$ where $I_{|_{\hat g_b}}$ is a (64$\times$64) crop  of the (centered and re-scaled) image $I\circ \hat{g_b}^{-1}$. We design a joint loss with multi-scale supervision for training pose and 3D scales.
\begin{align} \label{loss_azimuth}
& H_{P, Q}(\theta_j, \hat \theta_i) + \lambda d(\sigma_j, \hat \sigma_i)
\end{align}
where $Q$ is an approximation of $P(\theta | I, \hat g_b)$, both of which are Von Mises distributions over the discrete set $\{0,\dots,  359^o\}$. We use a cross-entropy loss for orientation estimation. At inference time, the MAP estimate $\hat{\theta}_i \in [0, 2\pi)$ is used as a vehicle's azimuth estimation. $P(\sigma|I,\hat g_b, \theta)$ is estimated jointly with azimuth in the same network using the $L^1$ distance $d(\sigma_j, \hat \sigma_i)$. Empirically, we found that imposing the orientation loss on the intermediate feature map and the size loss on the last layer generates better results than minimizing both losses on the same layer.

\subsubsection{Shape hypotheses.} The 3D morphable shape model is learnt using 2D landmarks via an EM-Gaussian method~\cite{torresani2004learning,kar2015category}. The hypothesis is that 3D object shapes are confined to a low-dimensional basis of the entire possible shape space. Therefore, the normalized 3D shape $S_{m} \in \mathbb{R}^{3K \times 1}$ of each vehicle instance can be factorized as the sum of the mean shape $\bar{S} \in \mathbb{R}^{3K \times 1}$ of this category deformed using a linear combination of $N$ basis shapes, $V_n \in \mathbb{R}^{3K \times 1}$. For each car, the orthographic projection constraint between its 3D shape $S_m$ and 2D landmarks $p_{k,m}$ is constructed as:
\begin{align} \label{em_morphable_shape}
& p_{k,m} = c_{m}R_{m}(P_{k,m}+t_{m}) + \zeta_{k,m}\\
& S_{m} = \bar{S} + \sum_{n=1}^{N}\alpha_{n,m}V_{n}\\
& \zeta_{k,m} \sim N(0,\sigma^{2}I_{2\times2}),\ \ \ \alpha_{n,m} \sim N(0,1),\ \ \ R_{m}^{T}R_{m} = I_{3\times3}
\end{align}
in which $c_{m} \in \mathbb{R}^{2 \times 2}$ is the scaling factor of the orthography (para-perspective projection). $R_{m} \in \mathbb{R}^{2 \times 3}$ and $t_{m} \in \mathbb{R}^{3 \times 1}$ are the orthographic rotation and translation of each object instance in the camera coordinate, respectively. $P_{k,m} \in \mathbb{R}^{3 \times 1}$ represents the $k_{th}$ 3D landmark in $S_{m}$.

\begin{table}
\centering
\resizebox{\columnwidth}{!}{

\begin{tabular}{|c|c|c|c|c|c|c|}
\hline
Method & 3DVP & 3DOP & Mono3D & \begin{tabular}[c]{@{}c@{}}GoogLenet\\ DeepMANTA\end{tabular} & \begin{tabular}[c]{@{}c@{}}VGG16\\ DeepMANTA\end{tabular} & Mono3D++\\ \hline
Type   & Mono & Stereo & Mono   & Mono                 & Mono             & Mono           \\ \hline
Time   & 40 s & 3 s    & 4.2 s  & 0.7 s                & 2 s              & \textbf{0.6 s} \\ \hline
\end{tabular}

}
\caption{Inference time comparisons against a paragon set of both monocular and stereo methods.}
\label{inference_time}
\end{table}

\begin{table*}
\centering
\resizebox{1 \textwidth}{!}{

\begin{tabular}{c|c|c|c|c|c|c|c|c|c|}
\cline{2-10}
                             & \multicolumn{3}{c|}{1 meter} & \multicolumn{3}{c|}{2 meters} & \multicolumn{3}{c|}{3 meters} \\ \hline
\multicolumn{1}{|c|}{Method} & Easy & Moderate & Hard & Easy & Moderate & Hard & Easy & Moderate & Hard \\ \hline
\multicolumn{1}{|c|}{3DVP} & 45.6 /$\ \ $-$\ \ \ $ & 34.3 /$\ \ $-$\ \ \ $ & 27.7 /$\ \ $-$\ \ \ $ & 65.7 /$\ \ $-$\ \ \ $ & 54.6 /$\ \ $-$\ \ \ $ & 45.6 /$\ \ $-$\ \ \ $ & $\ \ \ \ $-$\ \ $/$\ \ $-$\ \ \ $ & $\ \ \ \ $-$\ \ $/$\ \ $-$\ \ \ $ & $\ \ \ \ $-$\ \ $/$\ \ $-$\ \ \ $ \\ \hline
\multicolumn{1}{|c|}{SubCNN} & 39.3 /$\ \ $-$\ \ \ $ & 31.0 /$\ \ $-$\ \ \ $ & 26.0 /$\ \ $-$\ \ \ $ & 70.5 /$\ \ $-$\ \ \ $ & 56.2 /$\ \ $-$\ \ \ $ & 47.0 /$\ \ $-$\ \ \ $ & $\ \ \ \ $-$\ \ $/$\ \ $-$\ \ \ $ & $\ \ \ \ $-$\ \ $/$\ \ $-$\ \ \ $ & $\ \ \ \ $-$\ \ $/$\ \ $-$\ \ \ $ \\ \hline
\multicolumn{1}{|c|}{Mono3D} &$\ \ \ \ \ $-$\ \ $/ 46.0 &$\ \ \ $-$\ \ \ \ $/ 38.3 &$\ \ \ $-$\ \ \ \ $/ 34.0 &$\ \ \ $-$\ \ \ \ $/ 71.0 &$\ \ \ $-$\ \ \ \ $/ 59.9 &$\ \ \ $-$\ \ \ \ $/ 53.8 &$\ \ \ $-$\ \ \ \ $/ 80.3 &$\ \ \ \ \ $-$\ \ $/ 69.3 &$\ \ \ \ \ $-$\ \ $/ 62.7 \\ \hline
\multicolumn{1}{|c|}{GoogLenet DeepMANTA} & 70.9 / 65.7 & 58.1 / 53.8 & 49.0 / 47.2 & 90.1 / 89.3 & 77.0 / 75.9 & 66.1 / 67.3 & $\ \ \ \ $-$\ \ $/$\ \ $-$\ \ \ $ & $\ \ \ \ $-$\ \ $/$\ \ $-$\ \ \ $ & $\ \ \ \ $-$\ \ $/$\ \ $-$\ \ \ $ \\ \hline
\multicolumn{1}{|c|}{VGG16 DeepMANTA} & 66.9 / 69.7 & 53.2 / 54.4 & 44.4 / 47.8 & 88.3 / 91.0 & 74.3 / 76.4 & 63.6 / 67.8 & $\ \ \ \ $-$\ \ $/$\ \ $-$\ \ \ $ & $\ \ \ \ $-$\ \ $/$\ \ $-$\ \ \ $ & $\ \ \ \ $-$\ \ $/$\ \ $-$\ \ \ $ \\ \hline
\multicolumn{1}{|c|}{Mono3D++} & \textbf{80.6} / \textbf{80.2} & \textbf{67.7} / \textbf{65.1} & \textbf{56.0} / \textbf{54.6} & \textbf{93.3} / \textbf{92.7} & \textbf{83.0} / \textbf{80.8} & \textbf{71.8} / \textbf{70.5} & \textbf{95.0} / \textbf{95.4} & \textbf{86.7} / \textbf{85.4} & \textbf{76.2} / \textbf{75.9} \\ \hline
\end{tabular}

}
\caption{3D localization comparisons with monocular methods on KITTI val1/val2 by ALP of 1, 2 and 3 meters thresholds for 3D box center distance.}
\label{alp_mono}
\end{table*}

\subsubsection{Monocular depth.} $P(Z|I)$ is learned from stereo disparity, converted to depth using camera calibration. Similar to vehicle landmark detection, we use an hourglass network, with skip-connections, to predict per-pixel disparity~\cite{godard2017unsupervised}. The left view is input to the encoder, and the right view used as supervision for appearance matching at multiple output scales of the decoder. The total loss is accumulated across four output scales combining three terms. One measures the quality of image matching, one measures disparity smoothness, and the last measures left/right disparity consistency. Next we describe each term relative to the left view. Each term is replicated for the right view to enforce left-view consistency. Appearance is measured by 
\begin{align} \label{ap_term}
& \frac{1}{D}\sum_{a,b}\beta\frac{1-SSIM(I^l_{ab},\widetilde{I}^l_{ab})}{2}+(1-\beta)||I^l_{ab}-\widetilde{I}^l_{ab}||_1
\end{align}
which combines single-scale SSIM~\cite{wang2004image} and the $L^1$ distance between the input image $I^l$ and its reconstruction $\widetilde{I}^l$ obtained by warping $I^r$ using the disparity $d^r$ with a differentiable image sampler from the spatial transformer network~\cite{jaderberg2015spatial}. Disparity smoothness is measured by
\begin{align} \label{ds_term}
&  \frac{1}{D}\sum_{a,b}|\partial_{x}d_{ab}^{l}|e^{-||\partial_{x}I_{ab}^{l}||}+|\partial_{y}d_{ab}^{l}|e^{-||\partial_{y}I_{ab}^{l}||}
\end{align}
which contains an edge-aware term depending  $\partial I$~\cite{heise2013pm}. Finally, left/right consistency is measured using the $L^1$ norm.
\begin{align} \label{lr_term}
& \frac{1}{D}\sum_{a,b}\|d_{ab}^{l}-d_{ab-d_{ab}^{l}}^{r}\|_1.
\end{align}

\subsection{Implementation}
It takes about one week to train the 2D bounding box network, and two hours for the orientation/3D scale network on KITTI with 4 TITAN-X GPUs. The landmark detector is trained on Pascal3D. The training process for the monocular depth estimation network is unsupervised using KITTI stereo-pairs, which takes around 5 to 12 hours depending on the amount of data available. In theory, these deep networks could be unified into a single one and trained jointly, but this is beyond our scope here. Learning the morphable shape model takes about 2.5 minutes using 2D vehicle landmarks. At inference time, we use the Ceres solver~\cite{ceres-solver} to optimize the weighted loss~\eqref{joint_energy}. On average it converges in 1.5 milliseconds within about 15 iterations. Detailed timing comparisons are shown in Table~\ref{inference_time}.

\begin{table}
\centering
\resizebox{1 \columnwidth}{!}{

\begin{tabular}{cc|c|c|c|}
\cline{3-5}
                               &        & \multicolumn{3}{c|}{1 / 2 / 3 meters}                        \\ \hline
\multicolumn{1}{|c|}{Method}   & Type   & Easy               & Moderate           & Hard               \\ \hline
\multicolumn{1}{|c|}{Mono3D++} & Mono   & 80.2 / 92.7 / 95.4 & 65.1 / 80.8 / 85.4 & 54.6 / 70.5 / 75.9 \\ \hline
\multicolumn{1}{|c|}{3DOP}     & Stereo & 78.6 / 87.4 / 89.5 & 66.9 / 80.0 / 84.2 & 59.4 / 71.9 / 76.0 \\ \hline
\end{tabular}

}
\caption{3D localization comparisons with the state-of-the-art stereo method by ALP under 3D box center distance thresholds of 1, 2 and 3 meters. Note that our method only uses a single image for inference, while 3DOP needs stereo-pairs.}
\label{alp_stereo}
\end{table}

\section{Experiments}
We evaluate our method on the KITTI object detection benchmark. This dataset contains $7,481$ training images and $7,518$ test images. To facilitate comparison with competing approaches, we isolate a validation set from the training set according to the same protocol of~\cite{xiang2015data,xiang2017subcategory,chabot2017deep} called (train1, val1), and the same used by~\cite{3dopNIPS15,chen2016monocular,chabot2017deep} called (train2, val2). We report results using five evaluation metrics: three for 3D and two for 2D localization. For the former, we use average localization precision (ALP)~\cite{xiang2015data}, average precision based on 3D intersection-over-union (IoU), AP$_{3D}$, and bird's eye view based average localization precision, AP$_{loc}$~\cite{Geiger2012CVPR}. Although 2D localization is not our goal, as a sanity check we also measure average precision (AP) and average orientation similarity (AOS).

ALP is based on the distance between the center of the detected 3D boxes and the annotated ground truth. A 3D detection is correct if the distance is below a threshold. AP$_{3D}$ is computed from the IoU of 3D bounding boxes. AP$_{loc}$ is obtained by projecting the 3D bounding boxes to the ground plane (bird's eye view) and computing the 2D IoU with ground truth. Note that while ALP only measures distance between centers, both AP$_{3D}$ and AP$_{loc}$ jointly evaluate a vehicle's translation, orientation and 3D scales. AOS measures 2D orientation relative to ground truth.

\begin{figure*}
\centering
\resizebox{0.98 \textwidth}{!}{

\includegraphics[height=6.5cm]{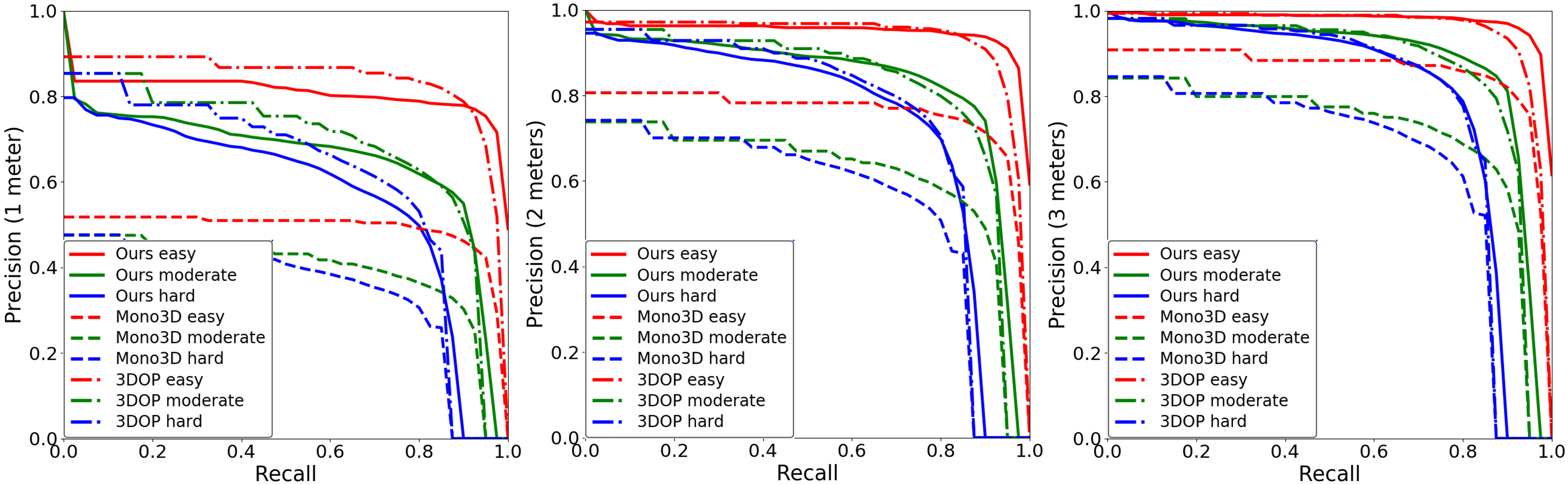}

}
\caption{Recall/3D localization precision curves for 1 meter (left), 2 meters (middle) and 3 meters (right) precision on KITTI val2. Solid lines are our results. Dashed lines are Mono3D. Dash-dot lines are 3DOP, which uses stereo-pairs at inference time.}
\label{pr_curves}
\end{figure*}

The paragon set for our experiments consists of 3DVP~\cite{xiang2015data}, SubCNN~\cite{xiang2017subcategory}, Mono3D~\cite{chen2016monocular}, 3DOP~\cite{3dopNIPS15} and DeepMANTA~\cite{chabot2017deep}. While 3DOP is the state-of-the-art stereo method, the rest are monocular. Among the monocular methods, DeepMANTA is the current state-of-the-art. Also related to our method for monocular 3D vehicle detection is Deep3DBox~\cite{mousavian20173d}, which however used different evaluation metrics from the ones above, thus preventing direct comparison. For object 2D orientation and bounding box evaluation, we also compare with Faster-RCNN~\cite{ren2015faster} as well as Deep3DBox.

\subsubsection{3D Localization.}
We use ALP with distance thresholds of $1$, $2$ and $3$ meters in Table~\ref{alp_mono} including both val1 and val2. Our method improves the state-of-the-art monocular method, DeepMANTA, by 10.5\% on average. Even though our method is monocular, we compare to the stereo method 3DOP using val2 in Table~\ref{alp_stereo}. Surprisingly, we outperform 3DOP on ``easy" and ``moderate'' cases and is comparable on ``hard" case. Detailed comparisons by precision/recall curves are shown in Fig.~\ref{pr_curves}. We outperform the monocular by large margins and is better than the stereo in some cases.

\subsubsection{3D Detection and Bird's Eye View Localization.}
We use AP$_{3D}$ with 3D IoU thresholds of $0.25$, $0.5$ and $0.7$, as well as AP$_{loc}$ with 2D IoU thresholds of $0.5$ and $0.7$. Table~\ref{3d_all} shows comparisons with both monocular and stereo methods using val2. Our method surpasses all monocular ones uniformly. Though the monocular setting is more challenging than the stereo one due to the lack of depth, our method still outperforms 3DOP by about 39\% to 61\% on AP$_{3D}$ with IoU threshold of $0.7$. Table~\ref{bv_all} shows comparison on AP$_{loc}$ with both monocular and stereo using val2. Again, our results surpass monocular ones uniformly. Even if compared with the stereo method, we gain around 21\% to 33\% relative improvement on AP$_{loc}$ under $0.7$ IoU.

\begin{table}
\centering
\resizebox{1 \columnwidth}{!}{


\begin{tabular}{cc|c|c|c|}
\cline{3-5}
                               &        & \multicolumn{3}{c|}{3D IoU 0.25 / 0.50 / 0.70}             \\ \hline
\multicolumn{1}{|c|}{Method}   & Type   & Easy               & Moderate          & Hard              \\ \hline
\multicolumn{1}{|c|}{Mono3D}   & Mono   & 62.9 / 25.2 / 2.5  & 48.2 / 18.2 / 2.3 & 42.7 / 15.5 / 2.3 \\ \hline
\multicolumn{1}{|c|}{Mono3D++} & Mono   & \textbf{71.9} / \textbf{42.0} / \textbf{10.6} & \textbf{59.1} / \textbf{29.8} / \textbf{7.9} & \textbf{50.5} / \textbf{24.2} / \textbf{5.7} \\ \hline \hline
\multicolumn{1}{|c|}{3DOP}     & Stereo & 85.5 / 46.0 / 6.6  & 68.8 / 34.6 / 5.1 & 64.1 / 30.1 / 4.1 \\ \hline
\end{tabular}

}
\caption{Comparisons on AP$_{3D}$ under different 3D IoU thresholds with both monocular and stereo methods.}
\label{3d_all}
\end{table}

\subsubsection{Ablation Studies.}
In Table~\ref{ablation_alp},~\ref{ablation_3d} and~\ref{ablation_bv} we use val1 and val2 with ALP, AP$_{3D}$ and AP$_{loc}$ to validate our joint modeling of the coarse and the fine-scaled 3D hypotheses, as well as task priors. ``v1'' indicates our inference scheme at initialization; ``v2'' only models the coarse geometric innovation and a ground plane constraint; ``v3'' adds the fine-scaled geometric innovation and vehicle shape priors. Best results are achieved by our overall model ``v4'', which further considers unsupervised monocular depth. Due to the page limit, extended comparisons over different threshold values are reported in the Supplementary Material.

\begin{table}
\centering
\resizebox{0.85 \columnwidth}{!}{


\begin{tabular}{cc|c|c|c|}
\cline{3-5}
                               &        & \multicolumn{3}{c|}{2D IoU  0.50 / 0.70}   \\ \hline
\multicolumn{1}{|c|}{Method}   & Type   & Easy        & Moderate    & Hard        \\ \hline
\multicolumn{1}{|c|}{Mono3D}   & Mono   & 30.5 / 5.2  & 22.4 / 5.2  & 19.2 / 4.1  \\ \hline
\multicolumn{1}{|c|}{Mono3D++} & Mono   & \textbf{46.7} / \textbf{16.7} & \textbf{34.3} / \textbf{11.5} & \textbf{28.1} / \textbf{10.1} \\ \hline \hline
\multicolumn{1}{|c|}{3DOP}     & Stereo & 55.0 / 12.6 & 41.3 / 9.5  & 34.6 / 7.6  \\ \hline
\end{tabular}

}
\caption{Comparisons on AP$_{loc}$ under different 2D IoU thresholds with both monocular and stereo methods.}
\label{bv_all}
\end{table}

\begin{table}
\centering
\resizebox{0.7 \columnwidth}{!}{

\begin{tabular}{c|c|c|c|c|c|c|c|c|c|c|}
\hline
\multicolumn{1}{|c|}{Method}   & \multicolumn{2}{c|}{Easy}                 & Moderate             & Hard \\ \hline
\multicolumn{1}{|c|}{v1} & \multicolumn{2}{c|}{13.6 / 16.9}          & 12.2 / 13.3          & 11.3 / 12.5 \\ \hline
\multicolumn{1}{|c|}{v2} & \multicolumn{2}{c|}{68.5 / 68.2}          & 58.3 / 57.5          & 50.8 / 47.6 \\ \hline
\multicolumn{1}{|c|}{v3} & \multicolumn{2}{c|}{76.1 / 73.2}          & 64.5 / 60.2          & 53.6 / 50.0 \\ \hline
\multicolumn{1}{|c|}{v4} & \multicolumn{2}{c|}{\textbf{80.6} / \textbf{80.2}} & \textbf{67.7} / \textbf{65.1} & \textbf{56.0} / \textbf{54.6} \\ \hline
\end{tabular}

}
\caption{Ablation studies on val1/val2 by ALP under 3D box center distance threshold of 1 meter.}
\label{ablation_alp}
\end{table}

\subsubsection{2D Detection and Orientation.}
As a sanity check, the 2D detection AP and AOS are also evaluated with monocular and stereo methods. Our estimation is on par with the state-of-the-art results. Detailed comparisons are included in the Supplementary Material.

\begin{table}
\centering
\resizebox{0.79 \columnwidth}{!}{

\begin{tabular}{c|c|c|c|c|c|c|c|c|c|c|}
\hline
\multicolumn{1}{|c|}{Method}   & \multicolumn{2}{c|}{Easy}                   & Moderate               & Hard \\ \hline
\multicolumn{1}{|c|}{v1} & \multicolumn{2}{c|}{10.50 / 11.50}          & 8.75 / 8.99            & 9.02 / 16.43 \\ \hline
\multicolumn{1}{|c|}{v2} & \multicolumn{2}{c|}{68.33 / 58.50}          & 55.00 / 49.96          & 49.17 / 45.09 \\ \hline
\multicolumn{1}{|c|}{v3} & \multicolumn{2}{c|}{71.39 / 66.59}          & 59.06 / 54.88          & 50.59 / 48.26 \\ \hline
\multicolumn{1}{|c|}{v4} & \multicolumn{2}{c|}{\textbf{79.45} / \textbf{71.86}} & \textbf{62.76} / \textbf{59.11} & \textbf{52.79} / \textbf{50.53} \\ \hline
\end{tabular}

}
\caption{Ablation studies on val1/val2 by AP$_{3D}$ with 3D IoU threshold of 0.25.}
\label{ablation_3d}
\end{table}

\begin{table}
\centering
\resizebox{0.80 \columnwidth}{!}{


\begin{tabular}{c|c|c|c|c|c|c|}
\hline
\multicolumn{1}{|c|}{Method}   & Easy                   & Moderate               & Hard \\ \hline
\multicolumn{1}{|c|}{v1} & 2.06 / 2.27            & 2.30 / 2.27            & 2.29 / 2.36 \\ \hline
\multicolumn{1}{|c|}{v2} & 37.27 / 30.18          & 27.48 / 24.82          & 23.67 / 21.49 \\ \hline
\multicolumn{1}{|c|}{v3} & 42.68 / 37.25          & 32.12 / 28.50          & 25.84 / 24.14 \\ \hline
\multicolumn{1}{|c|}{v4} & \textbf{50.50} / \textbf{46.68} & \textbf{36.85} / \textbf{34.32} & \textbf{29.05} / \textbf{28.13} \\ \hline
\end{tabular}

}
\caption{Ablation studies on val1/val2 by AP$_{loc}$ under 2D IoU threshold of 0.5.}
\label{ablation_bv}
\end{table}

\subsubsection{Qualitative Results.}
\begin{figure}
\centering
\resizebox{0.98 \columnwidth}{!}{

\includegraphics[height=6.5cm]{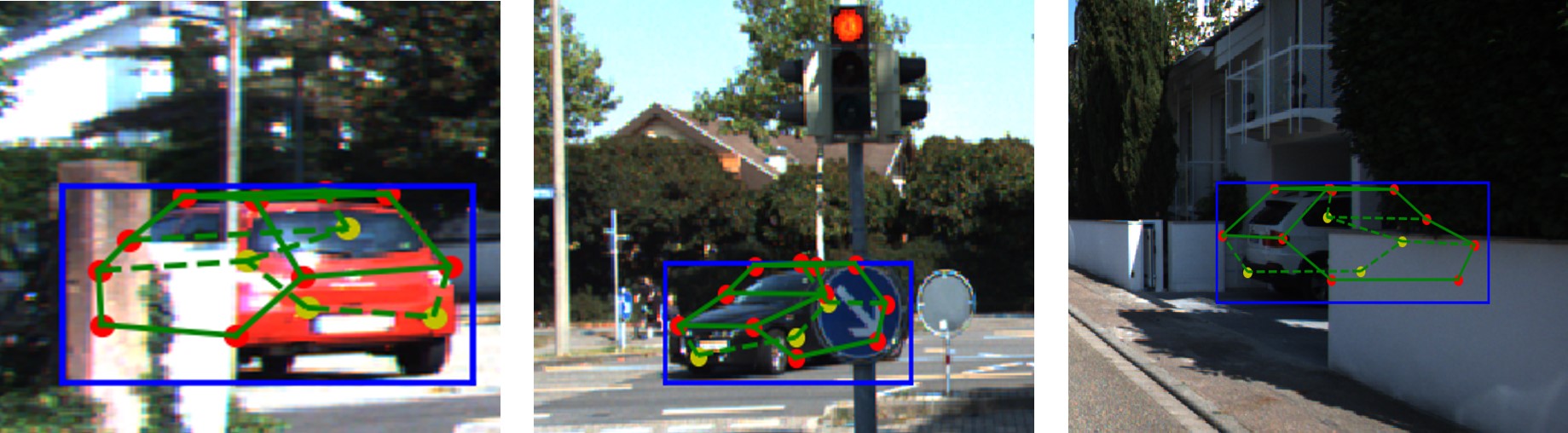}

}
\caption{Typical issues (e.g. partial occlusions, shadow, low resolution) that affect 2D vehicle landmark measurements.}
\label{landmark_occlusions}
\end{figure}

Fig.~\ref{visual_output} shows representative outputs of our method, including cars at different scales, 3D shapes, poses and occlusion patterns. Fig.~\ref{landmark_occlusions} illustrates typical issues addressed by jointly modeling a vehicle's coarse and fine-scaled 3D shape and pose representations. When vehicle landmarks suffer from partial occlusions, shadow or low resolution, we can still leverage 2D bounding boxes in order to enforce the two-scale geometric innovation constraints.

\subsubsection{Generality.}
Although not our focus here, chairs share similar constraints to vehicles like the two-scale 3D hypotheses innovation, a ground plane assumption, shape priors, etc. Thus to demonstrate the generality of our method, we also test on chairs and report results in the Supplementary Material.

\begin{figure}
\centering
\resizebox{0.98 \columnwidth}{!}{

\includegraphics[height=6.5cm]{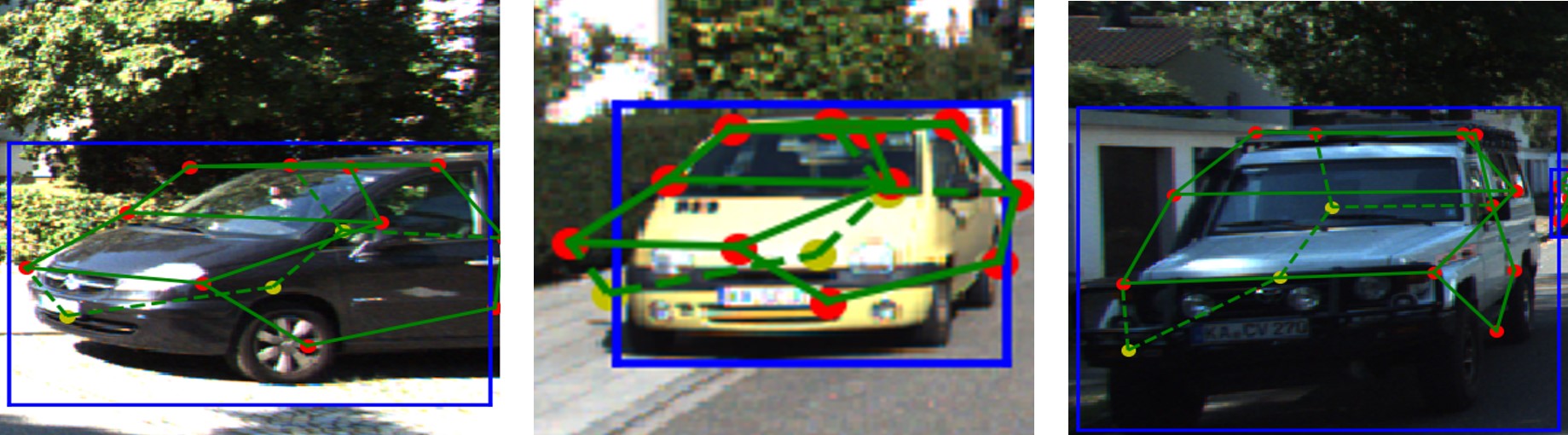}

}
\caption{Failure cases caused by field of view truncation, inaccurate orientation or 3D scale estimation.}
\label{visual_failures}
\end{figure}

\subsubsection{Failure Modes.}
In Fig.~\ref{visual_failures} we illustrate some failure cases, which include field of view truncation, causing the bounding box projection constraint $E_{\rm 2D3D}$ in the overall energy to enforce the 3D bounding box's 2D projection to be within the truncated 2D box measurement. Failures can also occur due to inaccurate orientation estimation, and under-representation in the training set (oversized SUV) which causes the normalized morphable wireframe model to be rescaled by incorrect 3D size estimation.

\section{Conclusion}
We have presented a method to infer vehicle pose and shape in 3D from a single RGB image that considers both the coarse and the fine-scaled 3D hypotheses, and multiple task priors, as reflected in an overall energy function~\eqref{joint_energy} for joint optimization. Our inference scheme leverages independence assumptions to decompose the posterior probability of pose and shape given an image into a number of factors. For each term we design a loss function that is initialized by output from deep network as shown in Fig.~\ref{pipeline_1}. Our method improves the state-of-the-art for monocular 3D vehicle detection under various evaluation settings.

\section*{Acknowledgement}
Research supported by ONR N00014-17-1-2072, N00014-13-1-034 and ARO W911NF-17-1-0304.

\bibliographystyle{aaai}
\bibliography{bibfile1}
\end{document}